\documentclass[runningheads]{llncs}
\usepackage[T1]{fontenc}
\usepackage{algorithm}
\usepackage{algorithmicx}
\usepackage{algpseudocode}
\usepackage{relsize}
\usepackage{scalefnt}
\usepackage{float}
\usepackage{afterpage} 
\usepackage{lipsum}
\usepackage{makecell}
\usepackage{amsmath}
\usepackage{amssymb}
\usepackage{amsfonts}
\usepackage{graphicx}
\usepackage{booktabs}
\usepackage{tabularx}
\usepackage[table]{xcolor}
\usepackage{multicol}
\usepackage{multirow}
\usepackage{hyperref}
\usepackage{cleveref}
\usepackage{caption}
\usepackage{array}
\usepackage{colortbl}
\usepackage{orcidlink}
\crefname{figure}{Fig.}{Figs.}  
\usepackage{graphicx}

\begin{document}
\title{SAR-NAS: Lightweight SAR Object Detection with Neural Architecture Search}
\titlerunning{SAR-NAS}

\author{Xinyi Yu\inst{1} \and
Zhiwei Lin\inst{1} \and
Yongtao Wang\inst{1}\correspondingauthor\orcidlink{0000-0003-1379-2206}}
\authorrunning{X. Yu et al.}
\newcommand\correspondingauthor{\thanks{Corresponding author.}}

%
\institute{Wangxuan Institute of Computer Technology, Peking University
\email{yuxinyi@stu.pku.edu.cn, \{zwlin,wangyongtao\}@pku.edu.cn}}

%
\maketitle              
\begin{abstract}
Synthetic Aperture Radar (SAR) object detection faces significant challenges from speckle noise, small target ambiguities, and on-board computational constraints. 
While existing approaches predominantly focus on SAR-specific architectural modifications, this paper explores the application of the existing lightweight object detector, \textit{i.e.}, YOLOv10, for SAR object detection and enhances its performance through Neural Architecture Search (NAS).
Specifically, we employ NAS to systematically optimize the network structure, especially focusing on the backbone architecture search.
%
By constructing an extensive search space and leveraging evolutionary search, our method identifies a favorable architecture that balances accuracy, parameter efficiency, and computational cost.
Notably, this work introduces NAS to SAR object detection for the first time.
The experimental results on the large-scale SARDet-100K dataset demonstrate that our optimized model outperforms existing SAR detection methods, achieving superior detection accuracy while maintaining lower computational overhead. 
We hope this work offers a novel perspective on leveraging NAS for real-world applications.

\end{abstract}
\section{Introduction}

Synthetic Aperture Radar (SAR) plays a crucial role in remote sensing applications, enabling all-weather and day-and-night environmental monitoring for land and maritime scenarios.
Unlike optical imaging, SAR imaging technology can penetrate cloud cover and deal with adverse weather conditions, making it an essential tool for large-scale Earth observation\cite{shao2021sar}. 
Nevertheless, significant challenges are faced in SAR object detection due to the inherent complexities of SAR imagery, including speckle noise, variable background clutter, and various object appearances across different imaging conditions.


Early methods predominantly utilize handcrafted feature engineering techniques, such as Constant False Alarm Rate\cite{liu2019cfar}, and Gabor filter-based texture analysis\cite{lu2015method}.
While these methods have achieved initial success in ideal scenarios, the dependence on handcrafted feature engineering leads to two limitations: (1) high false alarm rates and (2) poor generalization across diverse imaging conditions\cite{zhu2021deep}. 

\begin{figure}[tb]
  \centering
  \begin{minipage}{0.495\textwidth}
    \includegraphics[width=\linewidth]{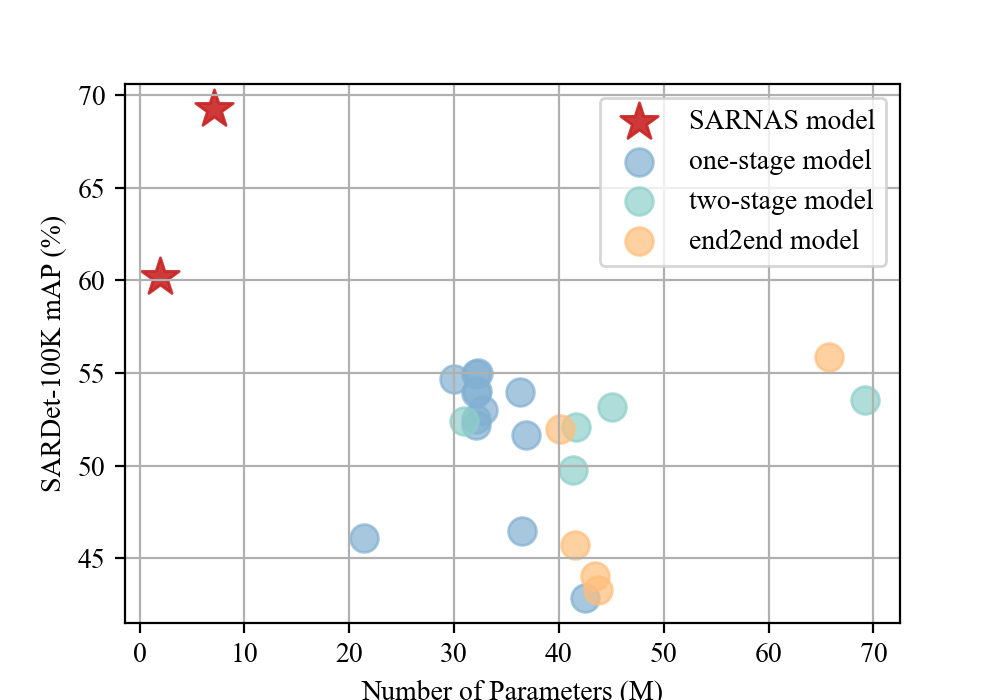}
  \end{minipage}
  \hfill
  \begin{minipage}{0.495\textwidth}
    \includegraphics[width=\linewidth]{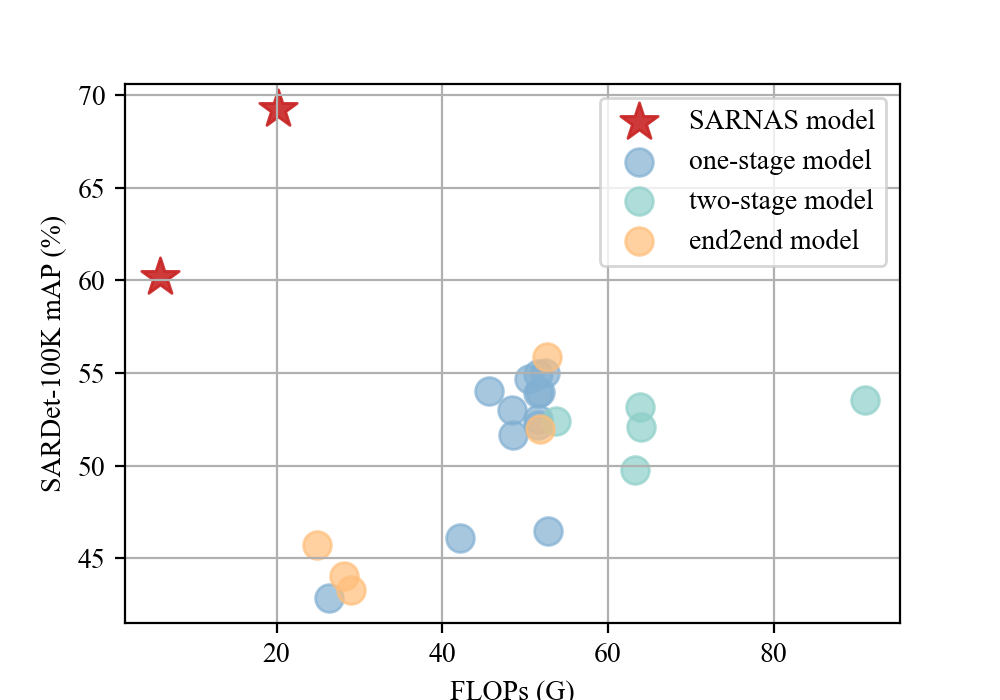}
  \end{minipage}
  \caption{\textbf{Comparison of the real-time object detectors on SARDet-100K dataset.} SAR-NAS achieves the best accuracy-efficiency trade-off than other models. }
  \label{fig:two_images}
\end{figure}

Recently, deep learning-based object detection methods have dramatically improved performance in SAR object detection. 
These approaches leverage convolutional neural networks (CNNs) and advanced object detection frameworks, including Faster R-CNN\cite{ren2015faster}, RetinaNet\cite{lin2017focal}, and the YOLO series\cite{ge2021yolox,bochkovskiy2020yolov4}, and achieve good detection performance for complex SAR imagery.
However, State-of-the-art (SOTA) methods\cite{dai2024denodet,wan2021afsar} still heavily depend on manual architecture design, which is time-consuming and computationally cost.
Additionally, many SOTA models are excessively large and difficult to deploy on mobile or resource-constrained devices for real-world applications.

To address these issues, this paper proposes SAR-NAS by leveraging the existing lightweight object detector, \textit{i.e.}, YOLOv10\cite{wang2024yolov10}, and Neural Architecture Search (NAS)\cite{guo2020single} to further enhance its performance while keeping efficiency. 
We selected YOLOv10 due to its high efficiency and modularity, which make it highly suitable for NAS integration.
Notably, its decoupled head and dynamic label assignment mechanism can effectively address common SAR challenges such as ambiguous object boundaries and class imbalance.
Specifically, we conduct extensive evaluations on the large-scale SARDet-100K dataset\cite{li2024sardet} to adapt YOLOv10 to SAR detection tasks, achieving promising detection results.
Then, based on the YOLOv10 baseline, we introduce NAS to explore more optimal architecture for SAR object detection. 
To the best of our knowledge, it is the first time to introduce NAS to the SAR object detection task.
Rather than proposing a brand-new NAS algorithm, our work positions itself as an early-stage feasibility study that demonstrates the potential of applying NAS to SAR detection, while paving the way for future designs incorporating SAR-specific priors.
The proposed NAS method mainly focuses on optimizing the channel dimensions of YOLOv10's backbone network. 
More concretely, we first construct a search space containing over four million candidate sub-network architectures. 
Afterward, by employing an evolutionary search strategy with hardware-aware constraints, including model parameters and computational cost (FLOPs), we can automatically identify an optimal architecture within less than 10 GPU-days on the V100 GPU.
 As shown in \cref{fig:two_images} the final searched model achieves a new state-of-the-art trade-off between accuracy, FLOPs, and parameter count, significantly outperforming existing SAR detection methods on SARDet-100K.

Our contributions can be summarized as follows:
\begin{itemize}
\item We introduce SAR-NAS, a lightweight SAR object detection method with NAS. To the best of our knowledge, it is the first time to introduce NAS to the SAR object detection task.

\item For the proposed NAS method, we mainly focus on channel configurations of the backbone network and construct an extensive NAS search space, containing over four million candidate sub-network architectures, to find the optimal architecture.

\item The extensive experiments on the large-scale SARDet-100K dataset show that SAR-NAS achieves a new state-of-the-art SAR detection performance. Moreover, SAR-NAS obtains the best accuracy-efficiency trade-off, facilitating model deployment in real-world applications.

\end{itemize}

\section{Related Work}
\subsection{Synthetic Aperture Radar Object Detection}
Traditional SAR object detection methods primarily rely on statistical models such as Constant False Alarm Rate\cite{liu2019cfar} and handcrafted feature-based approaches like Gabor filters\cite{lu2015method}.
These methods perform well in ideal conditions but struggle with high false alarm rates and poor adaptability in complex scenes with speckle noise and heterogeneous backgrounds.

Recently, the development of deep learning has significantly advanced SAR object detection by enabling automatic feature extraction. 
Early works adapt generic CNN architectures like VGG\cite{zhang2019high} and ResNet\cite{li2019sar} to SAR imagery, demonstrating notable performance improvement over traditional methods.  
Recent works\cite{liu2020sar} enhance SAR target detection by introducing frequency-domain transformations, which suppress speckle noise through convolutional bias calibration and strengthen high-frequency target signals.
Additionally, a dynamic pruning strategy\cite{wang2021boosting} improves computational efficiency by adaptively discarding irrelevant features.
However, existing methods still face critical limitations, \textit{i.e.}, high-performance detectors exhibit excessive computational redundancy, making edge deployment challenging.
Additionally, manual design optimizations require substantial expert intervention and computational resources, hindering rapid adaptation to evolving SAR detection demands.

In this paper, we introduce a lightweight and high-performance SAR object detection method with NAS to find the optimal architecture automatically.

\subsection{Neural Architecture Search for Object Detection}
Neural Architecture Search has emerged as a pivotal technique for automating network design. 

Early NAS methods relied on reinforcement learning and evolutionary algorithms to iteratively train isolated architectures, which requires high computational cost. 
To address this issue, approaches, such as Differentiable Architecture Search (DARTS) \cite{liu2018darts}, Single-Path One-Shot (SPOS)\cite{guo2020single}, and Once-for-All (OFA)\cite{cai2019once}, significantly reduce search cost by enabling weight sharing and gradient-based optimization.

In object detection, NAS has been successfully applied to optimize backbone networks and feature fusion mechanisms, including NAS-FPN\cite{ghiasi2019fpn} and SpineNet\cite{ao2021spinnet}, which achieve superior accuracy-efficiency trade-offs. 
Recent advances further integrate hardware-aware constraints.
EfficientDet\cite{tan2020efficientdet} employs compound scaling to balance accuracy and efficiency, while AutoFormer\cite{wu2021autoformer} adapts transformer-based architectures for mobile devices. 
However, NAS remains underexplored in SAR detection, where existing models still depend on manually designed CNNs. 

In this paper, at the first time, we introduce NAS for SAR object detection by designing an extensive search space and hardware-aware constraints.
\begin{figure}[tbp] 
\makebox[\textwidth][c]{\includegraphics[width=1.0\textwidth]{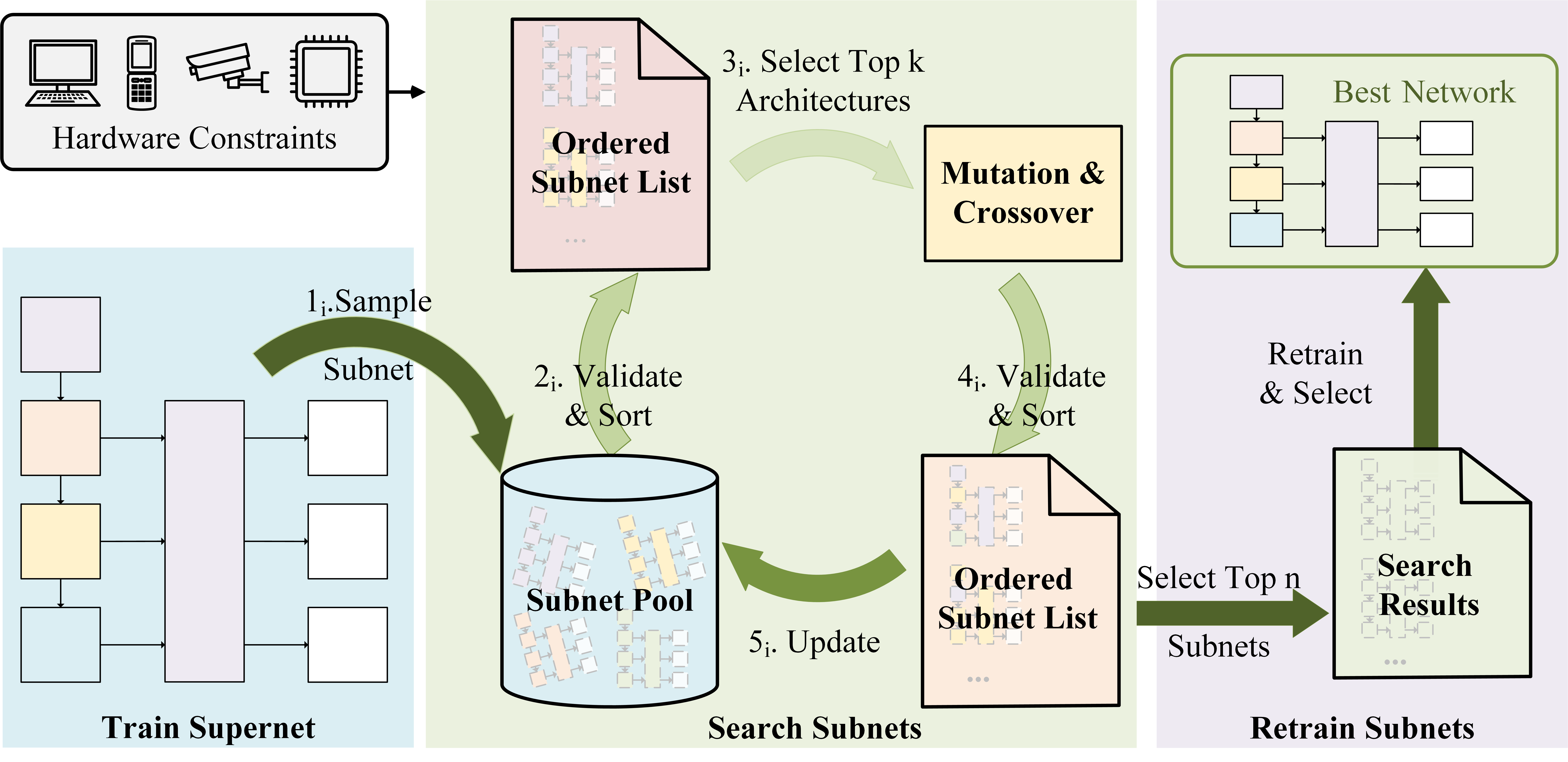}}  
  \centering
  \caption{\textbf{The pipeline of SAR-NAS.} 
  }
  \label{fig:pipeline}
\end{figure}

\section{Method}

In the following, we introduce the SAR-NAS pipeline in \cref{sec:sec31}. Then, we describe the design of the search space in \cref{sec:sec32}. Finally, we describe the search strategy in \cref{sec:sec33}.

\begin{figure}[tb]
\makebox[\textwidth][c]{\includegraphics[width=1.0\textwidth]{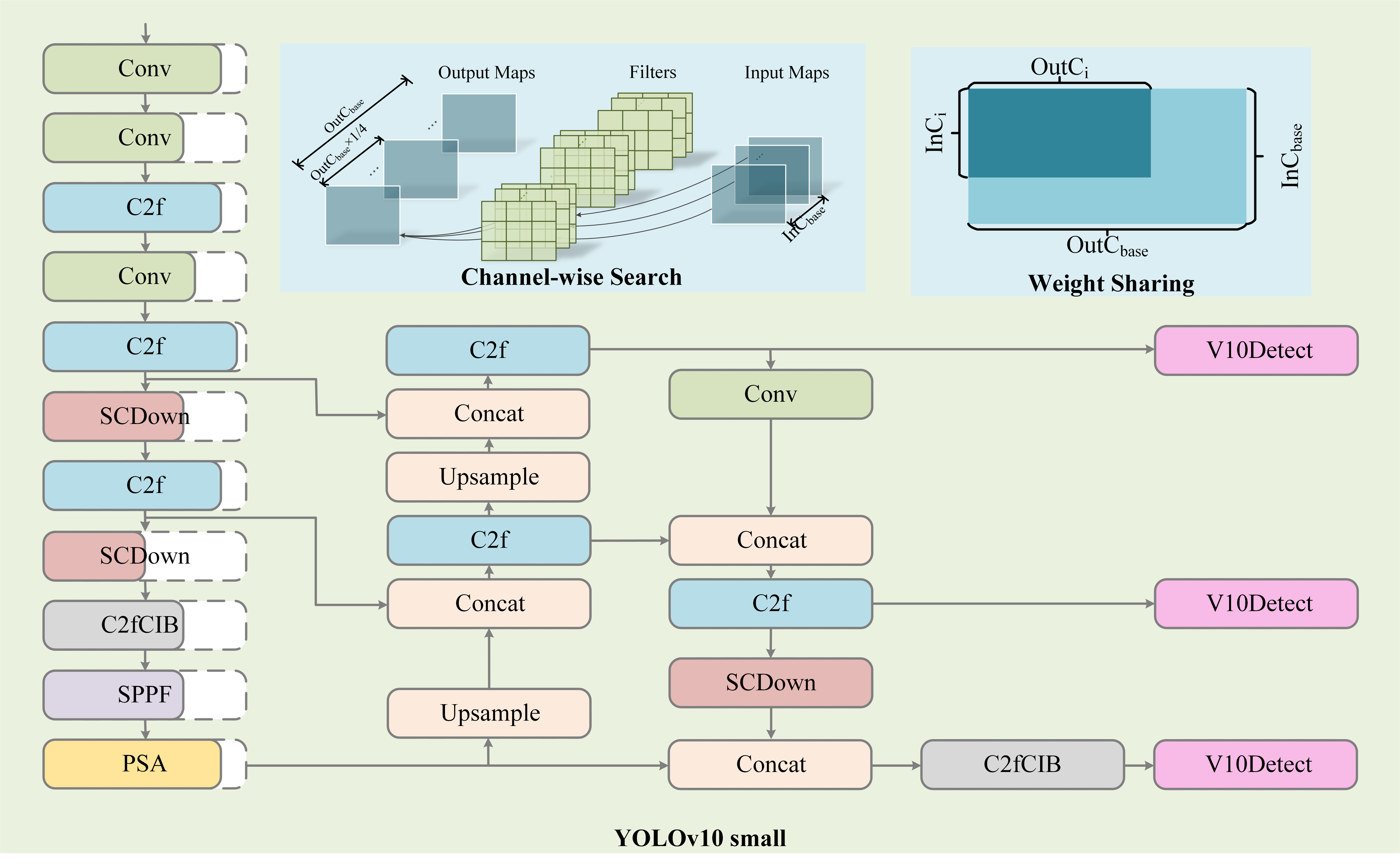}}  
  \centering
  \caption{\textbf{Search space design.}
  }
  \label{fig:search_space}
\end{figure}

\subsection{The Pipeline of SAR-NAS}
\label{sec:sec31}

Our SAR-NAS follows One-Shot NAS paradigm with hardware constraints. The search process leverages weight-sharing to inherit pre-trained supernet weights, ensuring efficient and accurate evaluation of candidate architectures. As illustrated in~\cref{fig:pipeline}, the pipeline consists of the following six steps.

\noindent\textbf{Supernet Training.} We first train a supernet that involves all possible sub-network configurations within the search space. During the supernet training process, we employ the sandwich training strategy, enabling accurate performance evaluation during the search phase.

\noindent\textbf{Architecture Sampling.} Candidate architectures are sampled from the search space and added to the architecture pool. Each sampled architecture follows a unique combination of channel configurations. 
During sampling, all candidate architectures adhere to predefined hardware constraints, ensuring that the selected model is computationally feasible for practical deployment.

\noindent\textbf{Weight Inheritance and Evaluation.} The sampled architectures in the pool inherit weights from the trained supernet and are evaluated on the validation dataset. These architectures are then sorted according to the mean Average Precision (mAP), resulting in an ordered architecture list.

\noindent\textbf{Mutation and Crossover.} The top-performing architectures from the ordered list are selected for mutation and crossover operations. This process will generate new candidate architectures inheriting from the top-performing ones.

\noindent\textbf{Update and Iteration.} The new architectures are evaluated and sorted. The updated list of architectures will replace the existing pool. This iterative process of sampling, evaluating, sorting and updating will continue for $T$ epochs.

\noindent\textbf{Retraining Top Architectures.} After $T$ epochs, the Top $N$ architectures under hardware constraints are selected and fully trained from scratch to obtain the final SAR-NAS model.

\renewcommand{\algorithmicrequire}{\textbf{Input:}}
\renewcommand{\algorithmicensure}{\textbf{Output:}}

 \floatname{algorithm}{Algorithm}
    \begin{algorithm}[H] 
    \setcounter{algorithm}{0}
        \caption{Evolutionary Search} \label{Algorithm:C4}
        \begin{algorithmic}[1] 
            \Require {Search space $S$, Trained supernet weights $W$, Validation dataset $D_{val}$, Population size $P$, Hardware constraints $C$, Epochs $T$, Mutation times $m$, Crossover times $m$, Probability $prob$, Selection top number $k$, Baseline architecture $F_0$.}

            \Ensure {$Top_{n}$ architectures in population $\mathcal{P}$.}

            \State Population $\mathcal{P} =[F_0]$, Accuracy records $\mathcal{R} =\varnothing$, $Top_{k}=\varnothing$.  \Comment{Initializing}

            \For{$i = 1,...,T$}
                 \While{$\left|\mathcal{P} \right| < P$} \Comment{Randomly sampling}
                \State Randomly generate neural architecture $arch$ under constraints $C$ from $\mathcal{S}$. 
                \State Inherit weights from $W$ for $arch$.
                \State Evaluate $acc$ = Validate($arch$, $W$, $D_val$).
                \State $\mathcal{P}$.append($arch$), $\mathcal{R}$.append($acc$).

                \EndWhile
                
                \State Sort $\mathcal{P}$ according to the accuracy scores in $\mathcal{R}$
                \State Update $Top{k}$ with top $k$ architectures in $\mathcal{P}$.
                \State $\mathcal{P}_{mutation} =\varnothing $, $\mathcal{P}_{crossover} =\varnothing $. 
                \For{$j = 1,...,m$} \Comment{Mutation}
                \State Generate architecture $arch_{m}$ via $Mutation(Top_{k})$ under constraints $C$.
                \State Inherit weights from $W$ for $arch$.
                \State Evaluate $acc$ = Validate($arch$, $W$, $D_val$).
                \State $\mathcal{P}_{mutation}$.append($arch_{m}$), $\mathcal{R}_{mutation}$.append($acc$).
            \EndFor
            \For{$j = 1,...,m$} \Comment{Crossover}
                \State Generate architecture $arch_{c}$ via $Crossover(Top_{k})$ under constraints $C$.
                \State Inherit weights from $W$ for $arch$.
                \State Evaluate $acc$ = Validate($arch$, $W$, $D_val$).
                \State $\mathcal{P}_{crossover}$.append($arch_{c}$),$\mathcal{R}_{crossover}$.append($acc$).
            \EndFor
                \State  $\mathcal{P} \gets \mathcal{P} \cup \mathcal{P}_{crossover} \cup \mathcal{P}_{mutation}$, $\mathcal{R} \gets \mathcal{R} \cup \mathcal{R}_{crossover} \cup \mathcal{R}_{mutation}$.
                \State Sort $\mathcal{P}$ according to the accuracy in $\mathcal{R}$, and remain the top $P$ architectures.

            \EndFor
            \State Select top $n$ architectures from $ \mathcal{P}$ as final results set $Top_{n}$.
            \State \Return{$Top_{n}$}.
        \end{algorithmic}
    \end{algorithm}

\subsection{Search Space and Supernet Training}
\label{sec:sec32}
\noindent\textbf{Search Space.}
The search space of SAR-NAS focuses on optimizing the backbone of YOLOv10, as shown in~\cref{fig:search_space}. 
Each layer’s channel configuration is selected from four possible scaling factors, \textit{i.e.}, 0.25, 0.5, 0.75, and 1.0, of the original channel width. 
This results in approximately 4,000,000 candidate architectures. 
The search space design follows the principles of Single Path One-Shot (SPOS) NAS, where all candidate architectures share a common supernet during training, reducing the overall computational cost of the search process.

\noindent\textbf{Supernet Training with Sandwich Rule}. 
To accelerate the training of weight-shared supernet, we adopt the sandwich rule\cite{yu2018slimmable} training strategy. 
This method ensures that the supernet is sufficiently trained and the inherited weights accurately reflect the performance of each sub-network. 
Specifically, during each training step, we sample and train the largest, smallest, and two randomly selected intermediate sub-networks. 
By leveraging this technique, we enable more sub-networks to participate in training and improve the reliability of performance evaluation during searching.

\subsection{Search Strategy}
\label{sec:sec33}

The overall evolutionary search process is detailed in~\cref{Algorithm:C4}. Specifically, given the predefined hardware constraints $C$, we randomly sample architectures from the search space, ensuring each architecture meets the hardware constraints. Then, each sampled architecture inherits weights from the trained supernet and is evaluated on the validation set.

During the mutation phase, each layer’s channel configuration is resampled with probability $p$ or remains unchanged with probability $1 - p$, ensuring a diverse set of architectures and preventing premature convergence. In the crossover phase, two parent architectures are selected, and for each layer, the configuration is inherited from parent-1 with probability $p$ or parent-2 with probability $1 - p$. This mechanism ensures that beneficial traits are propagated across generations, facilitating improved model performance.
The evolutionary process iterates for $T$ epochs until convergence.
Finally, the best architecture is selected for full training.

By incorporating weight-sharing and efficient evaluation, this approach significantly reduces the computational cost of NAS while obtaining high-performing SAR detection models.

\noindent\textbf{Hardware-Aware Constraints.} To ensure that the final architecture is practical for real-world deployment, hardware-aware constraints are incorporated during the search process. Specifically, the hardware-aware constraints: 1) \textit{Parameter Constraints}. The total number of parameters is limited to prevent excessive model size.
2) \textit{Computational Cost Constraints}. FLOPs are restricted to ensure that the model can inference in real-time.
By enforcing these constraints, SAR-NAS ensures that the final selected model achieves high accuracy and maintains a balance between detection performance and resource efficiency, making it suitable for deployment in practical SAR detection applications.

\section{Experiment}
In this section, we evaluate the effectiveness of our SAR-NAS framework through extensive experiments. In \cref{sec:sec41}, we detail the experimental settings, including dataset preparation, training configuration, and search parameters. In \cref{sec:sec42}, we compare our NAS-searched model with state-of-the-art SAR detection models on SARDet-100K, analyzing performance in terms of mAP, parameter count, and FLOPs. Finally, in \cref{sec:sec43}, we conduct an ablation study to verify the effectiveness of NAS in optimizing YOLOv10.

\begin{table*}[tbp]
  \centering
  \caption{Comparison with SOTA methods on the \textbf{SARDet-100K} dataset.}
  \label{tab:sardet}
  
  \scalebox{0.9}{
    \begin{tabular}{{l | p{2.5cm}<{\centering} | p{2.5cm}<{\centering} | p{2.5cm}<{\centering}}}
      \toprule
      Model & FLOPs(G)  & \#Params(M)  & mAP(\%) \\
      \midrule
      \multicolumn{4}{l}{\textit{One-stage}} \\
      \midrule
      FCOS~\cite{tian2019fcos}                  & 51.57    & 32.13                     & 52.52        \\
      GFL~\cite{li2020generalized}                & 52.36  & 32.27                       & \underline{55.01}         \\
      ATSS~\cite{zhang2020bridging}                & 51.57   & 32.13                     & 54.95        \\
      CenterNet~\cite{zhou2019objects}                & 51.55   & 32.12                      & 53.91         \\
      PAA~\cite{kim2020probabilistic}                & 51.57  & 32.13                       & 52.20         \\
      PVT-T~\cite{wang2021pyramid}                & 42.19   & 21.43                      & 46.10         \\
      RetinaNet~\cite{lin2017focal}                & 52.77 & 36.43                       & 46.48         \\
      TOOD~\cite{feng2021tood}                & 50.52 & 30.03                       & 54.65         \\
      DDOD~\cite{ACM2021DDOD}                & 45.58 & 32.21                       & 54.02         \\
      VFNet~\cite{zhang2021varifocalnet}                & 48.38 & 32.72                        & 53.01         \\
      AutoAssign~\cite{zhu2020autoassign}                & 51.83  & 36.26                       & 53.95        \\
      YOLOF~\cite{chen2021you}                & 26.32 & 42.46                        & 42.83        \\
      \midrule
      \multicolumn{4}{l}{\textit{Two-stage}} \\
      \midrule
      Faster R-CNN~\cite{ren2015faster}                & 63.2  & 41.37                       & 39.22         \\
      Cascade R-CNN~\cite{cai2019cascade}                & 90.99     & 69.17                    & 53.55         \\
      Dynamic R-CNN~\cite{zhang2020dynamic}                & 63.2      & 41.37                   & 49.75         \\
      Libra R-CNN~\cite{pang2019libra}                & 64.02       & 41.64                  & 52.09         \\
      ConvNeXt~\cite{CVPR2022ConvNeXt}                & 63.84          & 45.07              & 53.15         \\
      \midrule
      \multicolumn{4}{l}{\textit{End2end}} \\
      \midrule
      DETR~\cite{carion2020end}                & 24.94              & 41.56           & 45.73         \\
      Deformable DETR~\cite{zhu2020deformable}                & 51.78 & 40.10                        & 52.00         \\
      Conditional DETR~\cite{meng2021conditional}                & 28.09 & 43.45                         & 44.04         \\
      DenoDet~\cite{meng2021conditional}                & 52.69 & 65.78                          & 55.88        \\
      \midrule
      \rowcolor[rgb]{0.9,0.9,0.9}\textbf{SAR-NAS-N (Ours)}    & \textbf{5.96} & \textbf{1.97}            & \textbf{60.18}   \\
      \rowcolor[rgb]{0.9,0.9,0.9} \textbf{SAR-NAS-S (Ours)}    & \textbf{20.17} & \textbf{7.07}             & \textbf{69.27}   \\
      \bottomrule
    \end{tabular}
  }
  \vspace{\baselineskip}
\end{table*}

\subsection{Experiment Setting}
\label{sec:sec41}

We conduct our experiments on the SARDet-100K dataset, a large-scale benchmark for SAR object detection. 
%
SARDet-100K is the first large-scale, multi-class open-source SAR object detection benchmark. 
It merges and standardizes 10 prior SAR datasets, thus providing a highly diverse and representative testbed. 
We selected it as the primary evaluation dataset because of its scale, diversity, and representativeness. 
The supernet is trained for 500 epochs with a batch size of 256. 
We employ an evolutionary algorithm with a population size of $P$ = 50, the number of search iterations $T$ = 20, and mutation/crossover probabilities of $p$ = 0.1, mutation/crossover times $m$ = 25. The top 20 architectures are selected for mutation and crossover. After searching, the top 10 models are retrained from scratch, and the best-performing architecture is chosen as the final model.

\subsection{Main Results}
\label{sec:sec42}

As shown in \cref{fig:two_images} and \cref{tab:sardet}, we present a detailed comparison of the proposed SAR-NAS with state-of-the-art methods SAR object detection models on the SARDet-100K dataset.
The results indicate that SAR-NAS outperforms existing methods in terms of efficiency while maintaining high accuracy. 
Specifically, our SAR-NAS-S model achieves an mAP of 69.27\%, surpassing models such as Deformable DETR\cite{carion2020end}(52.00\%) and DenoDet\cite{meng2021conditional}(55.88\%), while significantly reducing computational cost. 
SAR-NAS-N, a more lightweight variant, achieves an mAP of 60.18\% with only 5.96GFLOPs and 1.97M parameters, which is considerably lower than other models, such as YOLOF\cite{chen2021you}(26.32GFLOPs, 42.46M parameters) and DETR\cite{carion2020end}(24.94GFLOPs, 41.56M parameters). 
These results confirm the efficiency and deployability of our searched models, making them highly suitable for real-world SAR target detection applications where computational resources are constrained.

\subsection{Ablation Study on NAS Effectiveness}
\label{sec:sec43}

To further validate the contribution of NAS in optimizing YOLOv10, we conduct an ablation study comparing the original YOLOv10 models with our SAR-NAS searched models. 
As shown in ~\cref{tab:comparison}, SAR-NAS demonstrates superior performance in both accuracy and computational efficiency.
Compared to YOLOv10-S\cite{wang2024yolov10}, which achieves an mAP of 68.84\% with 21.43GFLOPs and 7.22M parameters, SAR-NAS-S improves accuracy to 69.27\% while reducinGFLOPs to 20.17G and parameters to 7.07M, showing a significant enhancement in efficiency. 
Similarly, SAR-NAS-N achieves 60.18\% mAP with 5.96GFLOPs and 1.97M parameters, outperforming YOLOv10-N\cite{wang2024yolov10}(59.47\% mAP, 6.53GFLOPs, 2.27M parameters) in both accuracy and efficiency. 
These results confirm that NAS effectively discovers optimized architectures that balance accuracy and computational cost, making our approach a promising solution for real-world SAR target detection tasks.

\begin{table}[tbp]
\centering
\caption{Comparison of YOLOv10 and SAR-NAS searched models.}
\begin{tabular}{l|c|c|c|c|c|c}
\hline
Model & FLOPs(G) & $\Delta$ & \#Params(M) & $\Delta$ & mAP (\%) & $\Delta$ \\
\hline
YOLOv10-N\cite{wang2024yolov10} & 6.53 & - & 2.27 & - & 59.47 & - \\
\rowcolor[rgb]{0.9,0.9,0.9} SAR-NAS-N & 5.96 & -0.57 & 1.97 & -0.30 & 60.18 & +0.71 \\
YOLOv10-S\cite{wang2024yolov10} & 21.43 & - & 7.22 & - & 68.84 & - \\
\rowcolor[rgb]{0.9,0.9,0.9} SAR-NAS-S & 20.17 & -1.26 & 7.07 & -0.15 & 69.27 & +0.43 \\
\hline
\end{tabular}
\label{tab:comparison}
\end{table}

\noindent
\begin{table}[]
\centering
\caption{Channel Width Ratios of SAR-NAS Searched Models.}
\begin{tabular}{l |p{0.8cm} |p{0.8cm} |p{0.8cm}|p{0.8cm} |p{0.8cm} |p{0.8cm}|p{0.8cm} |p{0.8cm} |p{0.8cm}|p{0.8cm} |p{0.8cm} }
\hline
Model & \#1  & \#2 & \#3  & \#4  & \#5 & \#6 & \#7  & \#8 & \#9 & \#10  & \#11  \\
\hline
SAR-NAS-N  & 1.0× & 1.0× & 1.0× & 1.0× & 1.0× & 0.75× & 1.0× & 0.5× & 0.75× & 0.75× & 1.0×  \\ 
SAR-NAS-S  & 1.0× & 1.0× & 1.0× & 1.0× & 1.0× & 0.75× & 1.0× & 0.5× & 1.0× & 0.75× & 1.0×  \\
\hline
\end{tabular}
\label{tab:analysis}
\end{table}

\subsection{Architecture Trend Analysis}
To further explore the structural characteristics of the searched models, we analyze the distribution of channel configurations across different search results.
A key observation from our experiments is that the NAS-optimized architectures tend to adopt a consistent pattern in channel allocation to achieve efficiency while maintaining detection performance.

As shown in Table \ref{tab:analysis}, SAR-NAS preserves the original channel dimensions at key output layers feeding into the neck while reducing channels in other backbone layers, effectively minimizing model size and computation without compromising performance. Additionally, SAR-NAS tends to allocate more channels to early backbone layers, where feature extraction is most critical, while reducing channel dimensions in deeper layers to enhance efficiency. These patterns highlight SAR-NAS's ability to balance accuracy and efficiency, offering valuable insights for manual architecture design.

\section{Conclusion}
In this paper, we propose SAR-NAS, a Neural Architecture Search framework designed to optimize the YOLOv10 backbone for SAR object detection. Our approach automates architecture design, overcoming the inefficiencies of manual tuning by exploring a search space of over 4,000,000 candidate architectures. Leveraging a supernet-based training paradigm with sandwich rule training, SAR-NAS ensures reliable weight inheritance and accurate performance evaluation.
The evolutionary search process integrates hardware-aware constraints, enabling favorable trade-offs between accuracy and computational cost. Extensive experiments on SARDet-100K demonstrate that SAR-NAS achieves state-of-the-art performance and the best accuracy-efficiency trade-off. 
In addition, architecture trend analysis reveals NAS-optimized networks exhibit adaptive channel allocation. 
Our results offer a new perspective on SAR object detection model design for real-world applications.
In future work, we plan to explicitly incorporate SAR-specific priors into the NAS framework, such as texture-aware or speckle-resilient modules, and extend validation to more SAR datasets. 
We also envision combining NAS with model compression techniques to further enhance the practicality of SAR detection models under diverse hardware constraints.

\bibliographystyle{splncs04}
\bibliography{refrence.bib}

\end{document}